\documentclass[sigconf]{acmart}

\AtBeginDocument{%
  \providecommand\BibTeX{{%
    \normalfont B\kern-0.5em{\scshape i\kern-0.25em b}\kern-0.8em\TeX}}}

\acmConference[Conference’22]{Conference’22}{2022}{US}
\acmPrice{15.00}
\acmISBN{978-x-xxxx-xxxx-x/YY/MM}

\usepackage{bm}
\usepackage{amsmath}

\usepackage{comment}

\usepackage{subfigure}

\usepackage{multirow}

\begin{document}

\title{GUIM - General User and Item Embedding with Mixture of Representation in E-commerce}


\author{Chao Yang}
\authornote{These three authors contributed equally to this research.}
\affiliation{%
  \institution{Alibaba Group}
}
\email{xiuxin.yc@alibaba-inc.com}

\author{Ru He}
\authornotemark[1]
\affiliation{%
  \institution{Alibaba Group}
}
\email{ru.he@alibaba-inc.com}

\author{Fangquan Lin}
\authornotemark[1]
\affiliation{%
  \institution{Alibaba Group}
}
\email{fangquan.linfq@alibaba-inc.com}

\author{Suoyuan Song}
\affiliation{%
  \institution{Alibaba Group}
}
\email{suoyuan.ssy@alibaba-inc.com}

\author{Jingqiao Zhang}
\affiliation{%
  \institution{Alibaba Group}
}
\email{jingqiao.zhang@alibaba-inc.com}

\author{Cheng Yang}
\affiliation{%
  \institution{Alibaba Group}
}
\email{charis.yangc@alibaba-inc.com}

\renewcommand{\shortauthors}{Yang, He and  Lin, et al.}


\begin{abstract}
Our goal is to build general representation (embedding) for each user and each product item across Alibaba's businesses,
including Taobao and Tmall which are among the world's biggest e-commerce websites.
The representation of users and items has been playing a critical role in various downstream applications,
including recommendation system, search, marketing, demand forecasting and so on.
Inspired from the BERT model \citep{BERT_Devlin:arXiv2018} in natural language processing (NLP) domain,
we propose a GUIM (General User Item embedding with Mixture of representation) model to achieve the goal
with massive, structured, multi-modal
data including the interactions among hundreds of millions of users and items.
We utilize mixture of representation (MoR) as a novel representation form to model the diverse interests of each user.
In addition, we use the InfoNCE from contrastive learning \citep{CPC_Oord:arXiv2018} to avoid intractable computational costs due to the numerous size of item (token) vocabulary.
Finally, we propose a set of representative downstream tasks to serve as a standard benchmark to evaluate the quality of
the learned user and/or item embeddings, analogous to the GLUE benchmark \citep{GLUE_Wang_2018} in NLP domain.
Our experimental results in these downstream tasks clearly show the comparative value of embeddings  learned from our GUIM model.
\end{abstract}





\maketitle

\section{Introduction \label{sec_intro}}
In the last two decades, Alibaba has been developing a variety of businesses and platforms
in order to establish one of the world's largest e-commerce ecosystems to better serve customers' growing demand and diverse interests.
These platforms include the well-established large ones, such as Taobao (a C2C online shopping platform) and Tmall (a B2C platform),
as well as newly emerging platforms such as
Idle Fish (a platform for secondhand items) and Hema (a new-retail platform providing seamless in-store and online service).
All these platforms conduct many business tasks, including recommendation, search, marketing, item demand forecasting and so on.
The commonly used approach which intends to build an end-to-end model for each business task
becomes both labor-expensive and time-consuming,
as the number of business tasks is growing exponentially fast 
in the Alibaba ecosystem.
Furthermore, unlike these well-established platforms,
many newly emerging platforms do not have sufficient historical data
so that many powerful machine learning methods (especially deep learning methods)
cannot be used as end-to-end solutions.

To address these challenges,
we propose
to 
obtain general representation (embedding) for every user and every product item throughout the whole ecosystem,
which serves a common basis for various tasks across different platforms.
If such general representation throughout the whole ecosystem can be pre-trained,
it can be used for
various kinds of downstream business tasks across platforms through either feature-based approach or fine-tuning approach,
even if a downstream task does not have its own sufficient data.
In this paper, we propose a GUIM (General User and Item embedding with Mixture of representation) model
to generate such general user and item representation on the basis of massive, structured, multi-modal data involving the interactions among hundreds of millions of users and items across Alibaba platforms.

Our model is inspired by the BERT (Bidirectional Encoder Representations from Transformers) model \citep{BERT_Devlin:arXiv2018}, a recent milestone allowing self-supervised learning  in the natural language processing (NLP) domain.
The authors of the BERT propose the model to learn general language representation from large-scale unlabeled data in a pre-training phase, which can later be used for various downstream NLP tasks with each of their own labeled data in a fine-tuning phase.
Based on the Transformer model structure \citep{Transformer_Vaswani:NIPS2017},
the key component of the BERT is to introduce the "masked language
model" (MLM) pre-training task to predict the id of each randomly masked word token from both its left and its right context.
(The MLM task is also referred to as the Cloze task  \citep{Cloze_Taylor:Journalism1953}, but we will use the term MLM throughout our paper.)
Additionally,  BERT includes the "next sentence prediction" (NSP) pre-training task to predict whether two sentences in an input sequence are really a sentence-pair.
While the BERT model obtains the embedding of each word token from the textual input sequences,
our goal is to generate the embeddings of every item 
and every user from the interaction sequences between users and items along with additional multi-modal, structured side information.

Note that the embedding of each word token learned by the BERT is always a single vector of some fixed dimension (length),
which is a common representation strategy used in NLP. 
We notice that such a strategy, however, is often unable to fully describe the characteristics of a user in the e-commerce domain.
It is because a user often shows his/her diverse interests while interacting with various kinds of items,
especially when the user is active or stay in a platform for a long period of time,
as we observed from our user-item interaction data.
If an interest set of a user is uncommon among those of other users,
a single-vector representation is often unable to fully describe the characteristics of the user and leads to sub-optimal performance in the downstream tasks.  
Therefore, we propose to use a mixture-of-representation (MoR) strategy as a novel way for user representation.
In essence, we extend the original BERT model structure by introducing multiple "CLS"
special tokens so as to learn a set of vectors to represent each user.
Corresponding to these newly introduced CLS tokens,
we also introduce a new user-item matching task 
to further transform these CLS tokens to a set of embedding vectors that better represent diverse interests of a user on different kinds of products.

There is one additional challenge coming from the vocabulary of hundreds of millions of item tokens in our domain.
Such a large item vocabulary size makes it computationally intractable to enumerate over all the possible items from the full softmax operation, which has been commonly used in the BERT and many other NLP models.
To address this challenge, we use the InfoNCE method \citep{CPC_Oord:arXiv2018}  in our model,
by essentially sampling only some items from some noise distribution as negative cases.
The used InfoNCE method greatly reduces the computation costs but still keeps good learning performance,
by enjoying the good estimation property proven by \citet{NCE_and_NS_Ma:arXiv2018}.

The main contributions  of our paper are summarized as follows:

\begin{itemize}

\item We develop a model which uses large-scale structured, multi-modal data to learn all the user and item embeddings that are general enough for various downstream tasks across platforms in e-commerce, analogous to the BERT in the NLP domain.

\item Our model uses a mixture-of-representation (MoR) strategy as a novel way to learn diverse interests of a user, including multiple interests in both short term and long term.
We achieve this by introducing a set of CLS
tokens in the model as well as a new matching pre-training task.

\item
We uses the InfoNCE technique to achieve good performance while avoiding the intractable computational costs of the full softmax operation due to the vocabulary including hundreds of millions of item tokens.

\item We conduct experiments on different  tasks in e-commerce, and the results clearly show the value of
our learned embeddings in representative e-commerce scenarios.

\end{itemize}

\section{RELATED WORK \label{sec_related}}

\subsection{BERT Related Work}
As mentioned in Section \ref{sec_intro}, our model is inspired by  the BERT
model proposed by \citet{BERT_Devlin:arXiv2018} in the NLP domain.
The BERT model intends to
pre-learn general language representation  from large-scale unlabeled data, which can later be fine-tuned for various downstream NLP tasks.
The pre-trained BEERT model has empirically been shown to able achieve the state-of-the-art results on each of eleven NLP benchmark tasks after being fine-tuned with each corresponding labeled data.
Since the success of the BERT, many of its variants have merged in the NLP domain by adding or varying pre-training tasks \citep{RoBERTa_Liu_2019,SpanBERT_Joshi_2019,StructBERT_Wang_2019,XLNet_Yang_2019,UNILM_Dong_NIPS2019}.

While deep learning methods have been used in recommendation domain by many researchers
\citep{CDL_Wang:KDD2015,DNNYoutube_Covington:RecSys2016,RRN_Wu:WSDM2017,NCF_He:WWW2017,ACCM_Shi:CIKM2018},
the BERT4Rec, proposed by \citet{BERT4Rec_Sun:CIKM2019}, is 
a pioneer that
applies the core ideas of the BERT into sequential recommendation.
While mainly following the BERT model, the BERT4Rec changes each input from a sequence of word tokens (including two sentence segments) into a user's interaction sequence of item tokens in chronological order.
With each input sequence being a single sentence, the BERT4Rec eliminates the NSP task as well as segment embeddings.
In addition to input sequences with random masking similar to the original BERT,
the BERT4Rec generates extra input sequences by masking only the last item in the sequence in order to better conduct the sequential recommendation task (which is to predict the immediate next item).
As the authors clearly pointed out, the most critical difference between the original BERT and the BERT4Rec is that
BERT4Rec is an end-to-end model for sequential recommendation task,
instead of a model pre-trained to generate general representation for various downstream tasks.
The authors show the advantages of BERT4Rec over the previous sequential recommendation models through experimental results on four relatively small real-world datasets.

In contrast to the BERT4Rec, our GUIM intends to generate general embeddings for all the users and items throughout the whole e-commerce ecosystem by a pre-training procedure, which then can be used for various downstream tasks across platforms.
We also use a mixture-of-representation (MoR) strategy as a novel way to represent different aspects of a user
by introducing multiple CLS tokens and a new matching task to the model.

\subsection{Mixture of Softmaxes}
\citet{MoS_Yang:arXiv2017} use a matrix factorization framework to show a potential limitation called softmax bottleneck across the majority of softmax-based neural language models.
Roughly speaking, the softmax bottleneck explains that the expressiveness of softmax-based models is limited by the dimension of the word embeddings, regardless whether to use universal approximators to generate the embeddings or not.
The authors propose a language model called Mixture of Softmaxes (MoS) to mitigate this limitation.
The MoS model computes a set of softmax distributions and uses a weighted average of them for the probability estimation of next word.
The authors show that the MoS model improves the expressiveness over the softmax counterpart given the same embedding dimension.
Meanwhile, it states that the MoS model generalizes well and avoids the overfitting problem compared to non-parametric models or merely increasing the word embedding dimension.
Finally, the authors empirically show the advantages of the proposed MoS model on two language modeling datasets Penn Treebank (PTB) and WikiText-2 (WT2).

We note that in this paper we use the term of mixture of representation (MoR), instead of MoS, to emphasize our goal to generate general representation. In fact, while MoR and MoS is similar, we extend the definition of the weight used in the MoS mixture calculation to a more general form, as will be described in Section \ref{sec_proposed_approach}.
\\
\\

 \subsection{Noise Contrastive Estimation}

The BERT model and its variants utilize the full softmax loss function over
the whole vocabulary since its vocabulary size is usually less than 10,000 in the NLP domain.
In our scenario, however, the number of items is at least at the magnitude of 100 millions so that the full softmax becomes computationally impractical.

\citet{NCE_Gutmann:JMLR2012} propose a principal called the Noise Contrastive Estimation (NCE) which avoids computing the full softmax over the whole vocabulary,
while achieving desirable theoretical properties such as asymptotic convergence to the MLE (maximum likelihood estimation) of the observed data. It introduces a noise distribution and set up a surrogate binary classification task to discriminate between (positive) samples from observed data distribution and (negative) samples from noise distribution.
The NCE method has been used by \citet{Skip_gram_Mikolov:NIPS2013} to improve the efficiency of their previously proposed Skip-gram Model \citep{CBOW_SG_Mikolov:ICLR2013}.


Following the same principal, \citet{CPC_Oord:arXiv2018} propose the so-called InfoNCE through an alternative surrogate multi-class classification to identify a positive sample against multiple negative samples.
They 
also shed light on
the theoretical connection between the InfoNCE and the mutual information.

We note that the InfoNCE has actually been proposed and studied under different names in earlier literature.
\citet{Jozefowicz:2016exploring} compared the InfoNCE (called IS in their paper) with the NCE on a 1B word benchmark dataset (with a vocabulary of 793,471 words),
and the InfoNCE shows better performance and faster convergence up to 50 epochs in their comparison.
\citet{NCE_and_NS_Ma:arXiv2018} provide a comprehensive, comparative theoretical study between the NCE and the InfoNCE, named binary and ranking objectives in their paper respectively.
They prove that the NCE has a stronger underlying assumption than the InfoNCE for self-normalization which might not be taken for granted in reality.
Their experiments show better performance of the InfoNCE when a certain regularizer is introduced.

Based on these previous studies on the InfoNCE, we have adopted the InfoNCE in our GUIM to avoid the intractable computational costs from the full softmax operation in our domain.

\section{PROPOSED APPROACH \label{sec_proposed_approach}}

In this section, we describe the proposed GUIM (general user and item embedding with mixture of representation) model, including the model structure, the mixture of representation, and the InfoNCE loss.

\subsection{Notations}

\begin{figure*}[h]
  \centering
  \includegraphics[width=\linewidth]{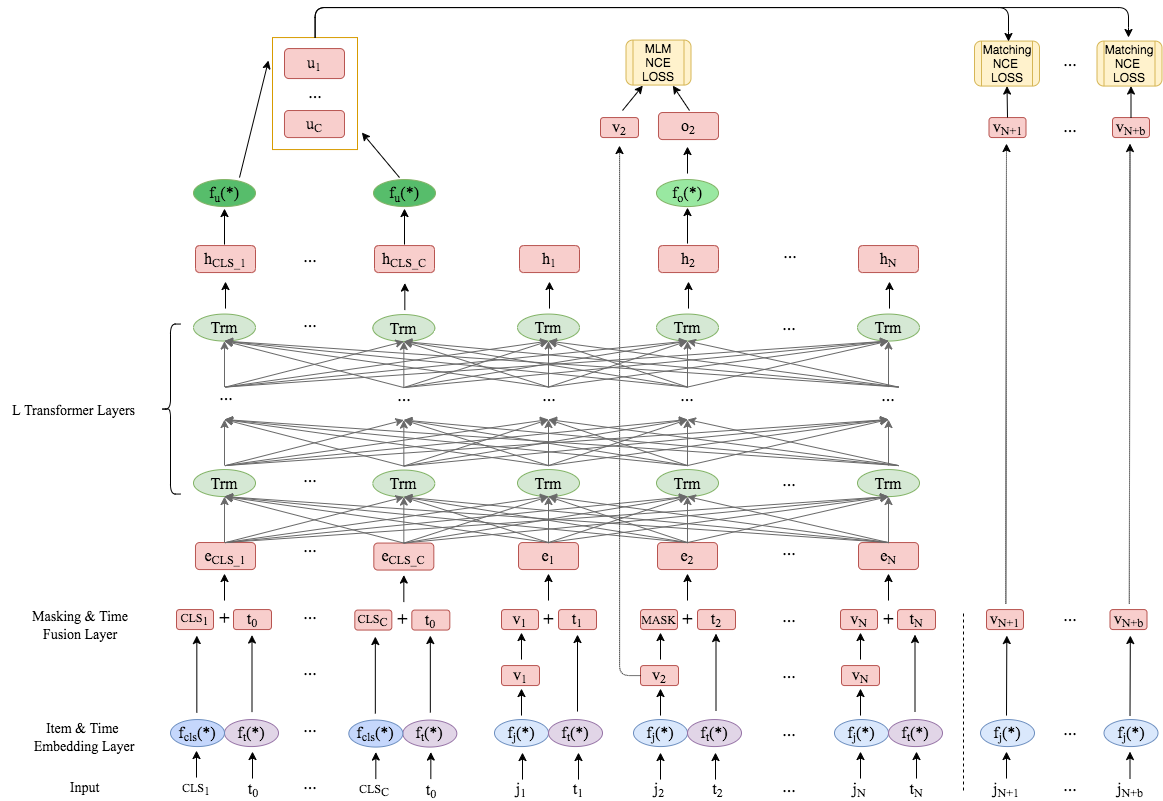}
  \caption{GUIM Model Structure}
  \label{fig_GUIM_model}
\end{figure*}

We use index set
$\mathcal{I}  = \{1, 2, \cdots, |\mathcal{I}| \}$
to denote a set of all the users
and
$\mathcal{J}  = \{1, 2, \cdots, |\mathcal{J}| \}$
to denote a set of all the items.
Let the interaction sequence of user $i$ in the period $[T-D_1, T)$ be denoted by
\begin{equation}\label{eq_interaction_sequence}
S_{[T-D_1, T)}^{(i)} = \langle a_1^{(i)}, a_2^{(i)}, \cdots, a_n^{(i)}, \cdots,a_{N^{(i)}}^{(i)} \rangle,
\end{equation}
where $a_n^{(i)}$ corresponds to the $n$-th interaction of user $i$ in the period of length $D_1$ before a given cutoff time $T$.
$a_n^{(i)}$ represents a tuple of information entities associated with the interaction,
such as the interacted item information, the action time, the action type and the action location.
In this paper, we focus on one action type "purchase" (so that an interaction sequence of a user is essentially his/her purchase sequence),
by noting that other action types such as clicking, browsing and searching can also be seamlessly added into our model (by introducing action-type embeddings).
In this paper, we also focus on the item information and the action time,
and denote $a_n^{(i)} := (j_n^{(i)}, t_n^{(i)})$,
meaning that user $i$ interacted with (essentially purchased) item $j_n$ at time ${t_n}$. 
The item information contains item id, item's category id as well as the (textual) title description of the item.
Also note that the subscript $N^{(i)}$ in Eq. (\ref{eq_interaction_sequence}) denotes the number of interactions of user $i$ in the time period from $T - D_1$ to $T$.
Similarly, we let
$  S_{[T, T + D_2)}^{(i)} $  $= \langle a_{N^{(i)} + 1}^{(i)},   a_{N^{(i)} + 2}^{(i)},
\cdots, a_{N^{(i)} + b^{(i)}}^{(i)} \rangle$ denote the interaction sequence of user $i$  in chronological order within the time period from $T$ to $T + D_2$,
where $b^{(i)}$ is the number of interactions of user $i$ in the period.
Essentially, we have partitioned the whole interaction sequence of user $i$ in the time period $[T-D_1, T+D_2)$ into two sub-sequences $  S_{[T-D_1, T)}^{(i)} $ and $  S_{[T, T + D_2)}^{(i)} $ according to the cutoff time $T$.
As described soon, we will assign one pre-training task (MLM task) in the first sub-sequence $  S_{[T-D_1, T)}^{(i)} $
and the other pre-training task (matching task) in the second sub-sequence  $  S_{[T, T + D_2)}^{(i)} $.

Remember that our goal is to learn an embedding of each user and an embedding of each item.
We represent the target embedding for item $j$  as vector $\bm{v_j} \in \mathbb{R}^d$, a vector of length $d$.
We propose to represent the target embedding for user $i$ as
$\bm{U^{(i)}} = \{ \bm{u^{(i)}_{1}},  \cdots, \bm{u^{(i)}_{C}} \}$,
a set of $C$ vectors of dimension $d$.
As shown later,  using a set of vectors to represent a user has its advantages over a single vector in
modeling diverse interests of the user in different items.

\subsection{Model Structure}

The structure of our GUIM model is demonstrated in Figure \ref{fig_GUIM_model}.
In the figure, we use each red rectangle to represent a vector of length $d$ and each oval to represent a function.
The data input to the model is the concatenation of $  S_{[T-D_1, T)}^{(i)} $ and $  S_{[T,T+D_2)}^{(i)} $ defined above,
i.e., the whole interaction sequence of each user $i$ in the period $[T-D_1, T+D_2)$.
We generate a CLS head sequence, a sequence of $C$ components
$\langle (CLS_1, t_0), (CLS_2, t_0), \cdots,  (CLS_C, t_0) \rangle$,
and concatenate it ahead of the whole interaction sequence of each user $i$.
$CLS_c$ for $c \in \{1, \cdots, C \}$ are $C$ special tokens, and
$t_0$ is a special time stamp indicating that it is before the starting point $T-D_1$.
Note that the CLS head sequence is the same for the interaction sequence of each user.
As shown on the bottom of Figure \ref{fig_GUIM_model}, the concatenated input sequence is the final input to our GUIM model.
For notational convenience,  in the following description we will omit the superscript $(i)$
unless the context of user $i$ needs to be emphasized.

The first layer of our GUIM model is  "Item \& Time  Embedding Layer".
In this layer, the function $f_j()$ is responsible for generating the item embedding $\bm{v_j}$ of dimension $d$
for each item $j$ in the input sequence.
Recall that the information of each item includes item id, item's category id and its (textual) title description.
Based on the category id of a given item,
the corresponding category-id embedding of dimension $d_c$ can be looked up from a category-id embedding table inside $f_j()$.
Similarly, given an item id, the item-id embedding of dimension $d_i$ can be retrieved from an item-id embedding table inside $f_j()$.
Inside the item-id embedding table, however, we only include the ids of top items (i.e., most popularly sold items)
and map all the other item ids into one additional common 
id.
By this way, we avoid exploding the number of parameters in the associated item-id embedding table due to billions of item ids in our system.\footnote{The total number of category ids in our system is only about 10k so that the size of category-id embedding table is not an issue.}
In practice, we have empirically studied
the impact of $X$, the number of top item ids included in the embedding table.
From our experiments on $X \in \{10k, 20k, 50k, 100k, 200k, 500k\}$,
we choose $X = 200k$ where the marginal benefit becomes too small to justify a further increase in $X$.
As for the title description of an given item (i.e., a sequence of words describing the item),
we get the word embedding of dimension $d_w$ from a word-level embedding table (with vocabulary size 530k) for each (textual) word in the description,
and then average across all the words.
Finally, we concatenate the category-id embedding, the item-id embedding and the average word embedding,
and feed the concatenated vector of dimension $d_c + d_i + d_w$ into a one-layer fully connected feed-forward network (FFN)
to obtain the final embedding of dimension $d$ for the given item.
The one-layer FFN consists of a linear transformation followed by a Gaussian Error Linear Unit (GELU) activation function \citep{GELU_Hendrycks:arXiv2016}.
Note that we use a simple averaging operation across words as well as a one-layer FFN for the final projection,
instead of more complicated networks such as transformers,
for the purpose of efficient inference in production.

In addition to $f_j()$, we use function $f_{cls}()$ to lookup an embedding of dimension $d$ for special token $CLS_c$ for $c \in \{1, \cdots, C\}$ from its corresponding embedding table.
We also use $f_{t}()$ to generate a $d$-dimensional  time embedding for a given time stamp $t_n$ by converting $t_n$ to a categorical value at a daily granularity and then looking up the corresponding time embedding table.\footnote{Other time granularities such as hour can also be used in our GUIM model by adjusting the number of embeddings in the time embedding table.}
Note that though our time embedding is in analogy with the position embedding used in the original BERT,
several items could have the same time embedding if they are purchased within the same day by a user.

The second layer of our model is "Masking \& Time Fusion Layer".
The logic of this layer is similar to that in the original BERT.
In the first step, with some specified probability it randomly chooses some items from $j_1$to $j_N$ in the sequence for the masking purpose, that is, replaces the embedding of each chosen item with the embedding of a special token "MASK".
In our model, we use a default setting of 15\% masking probability, just as the original BERT.
In the second step,
for each embedding of $CLS_c$ or an item before the cutoff time $T$,
it adds with the corresponding time embedding to generate the final embedding
which will be fed to the consequent Transformer layers.

On the top of Masking \& Time Fusion Layer,
$L$ transformer layers are incorporated in our model.
Since these transformer layers are the same as the ones in the original BERT,
we omit the corresponding description. 
The output generated from the transformer layers includes a $d$-dimensional embedding $\bm{h_n} $ for each item $j_n \in \{j_1, \cdots, j_N\}$
as well as a $d$-dimensional embedding $\bm{h_{CLS_c}}$ for each $CLS_c$.
Finally, for each $\bm{h_n}$,
the function $f_o()$ applies
its one-layer FFN
to generate an embedding $\bm{o_n} \in \mathbb{R}^d$.
Similarly, for each $\bm{h_{CLS_c}}$,
the function $f_u()$ applies
its one-layer FFN
to generate an embedding $\bm{u_c} \in \mathbb{R}^d$.
The resulting set $\{\bm{u_1}, \cdots, \bm{u_C} \}$ is the final user representation.

We include two pre-training tasks in our model:
one is the MLM task, the other is the matching task.
Since the MLM task is similar to that in the original BERT (except that
the InfoNCE is used to avoid the computation-intractable full softmax operation),
we focus on describing our proposed matching task which is
based on user $i$ and his/her interacted items in the time period $[T, T+D_2)$.

\subsection{Mixture of Representation \& InfoNCE}

In order to conduct the matching task, we design a score function $s^{\bm{\theta}}(i, j)$
for the mixture of representation of a user.
$s^{\bm{\theta}}(i, j)$ is to measure the association strength between user $i$ and item $j$,
where vector $\bm{\theta}$ are the correspondingly parameters in the GUIM model.
The score function is formulated as follows:
\begin{equation}\label{eq_preference_score_general}
s^{\bm{\theta}}(i, j) = \alpha \sum_{c=1}^C  \{ \pi_{c|i, j} cos(  \bm{u^{(i)}_{c}},  \bm{v_j} ) \}
\end{equation}
with the constraint $ \sum_{c=1}^C  \pi_{c|i, j} = 1,$ where the weight $\pi_{c|i, j}$ can be computed by some attention mechanism involving both user $i$ and item $j$, and $\alpha > 0$ is a constant which essentially acts as a scaling coefficient
so that the value of $s^{\bm{\theta}}(i, j)$ resides in the interval $[-\alpha, \alpha]$.\footnote{Empirically, we find the value $20$ is a good value for hyper-parameter $\alpha$ and applies this value in our experiments.}
Eq. (\ref{eq_preference_score_general}) is very expressive  because the value $\pi_{c|i, j}$ depends on each pair of user $i$ and item $j$.
However,
because many of our downstream tasks involve
searching the items with top association values to a given user from hundreds of millions items,
using Eq. (\ref{eq_preference_score_general}) in these downstream tasks is very computationally demanding.
To address this issue,
in these downstream tasks we intend to use the vector indexing engine already built inside Alibaba,
which is able to efficiently find the items most associated with a given user 
as long as input user embedding and item embedding are vectors of the same dimension.
Therefore, we deploy an alternative score function as follows:
\begin{equation}\label{eq_preference_score}
s^{\bm{\theta}}(i, j) = \alpha \max_{c \in  \{1, \cdots, C\} }  \{ cos(  \bm{u^{(i)}_{c}},  \bm{v_j} ) \},
\end{equation}
which can be regarded as a special case of Eq. (\ref{eq_preference_score_general}) by requiring that
$\pi_{c|i, j} = 1$ for one particular $c$.
With this score function,
we can call the vector indexing service in parallel by using each obtained $\bm{u^{(i)}_{c}}$ as the user vector for user $i$ in these downstream tasks.

As for the InfoNCE,
for each item $j \in \{j_{N+1}, j_{N+1},  \cdots, j_{N+b} \}$ in the period $[T, T+D_2)$,
we independently 
perform a matching task using the InfoNCE.
More specifically,
for each $j \in \{j_{N+1}, j_{N+1},  \cdots, $ $j_{N+b} \}$,
we sample $K$ negative cases from some noise distribution $q_J(j)$,
and combine them with the positive case $j$,
and finally infer which of the $K + 1$ cases is positive.\footnote{The MLM task
is also conducted using the InfoNCE with $K$ negative samples in a similar fashion for each masked item to avoid the computationally intractable full softamx operation.}
In our model,
we actually sample $K$ negative items from the interaction sequences of other users
which reside in the same mini-batch as
the sequence of user $i$.\footnote{In our model training we set $K$ to be $31$,
because our batch size is $32$ so that at least $31$ negative items are available for each user.}
Because we have randomly shuffled the input data, the corresponding noise distribution $q_J(j)$ is roughly equal to
the purchase frequency of item $j$.
Without loss of generality,  we label the $K$ negative samples as $j(1), j(2), \cdots, j(K)$,
and label the positive case $j$ as $j(0)$.

We apply our score function in the InfoNCE as follows.
We use our score function $s^{\bm{\theta}}(i, j)$  through the softmax function to get the 
probability
$p_{A | I J^{K+1}} (a | i, j(0), j(1), \cdots, j(K))$ for random variable $A$ so that:
\begin{equation}\label{eq_param_posterior}
p^{\bm{\theta}}_{A | I J^{K+1}} (a | i, j(0), j(1), \cdots, j(K))
= \frac{ exp(s^{\bm{\theta}}(i, j(a))) }{\sum_{k = 0}^K exp(s^{\bm{\theta}}(i, j(k))) },
\end{equation}
where random variable $A$ has a multinomial distribution whose domain is $\{0, 1, \cdots, K \}$.
Given our labelled data,
our overall matching loss is the summation of negative log likelihood for each user $i$ and his corresponding item $j \in \{j_{N+1}, j_{N+1},  \cdots, j_{N+b} \}$,
which is  $- \sum_{I,J} \log p^{\bm{\theta}}_{A | I J^{K+1}} (a=0 | i, j(0), j(1), \cdots, j(K)) $.\footnote{The overall loss of
our GUIM model is the summation of the overall matching loss and the overall MLM loss.}

As a final note,
because of our MoR strategy along with our score function listed in Eq. (\ref{eq_preference_score}),
our softmax listed in
Eq. (\ref{eq_param_posterior}) is
also able to
break the softmax bottleneck mentioned by \citet{MoS_Yang:arXiv2017}.

\section{EXPERIMENTAL RESULTS \label{sec_evaluation}}
We use a set of representative datasets collected from Alibaba's  e-commerce platforms to evaluate the performance of our proposed GUIM model.
To the best of our knowledge, there is no large-scale open-source benchmark dataset with multi-modal side information
which can be used to train or evaluate the general user and item representation.
Therefore, we use the real 
records from Taobao and Tmall,
two largest e-commerce websites at Alibaba,
for the training and evaluation purpose.

\subsection{Pre-training Data \& Model}


\begin{table*}
  \caption{Statistics of Number of Purchases per User in Pre-training Period}
  \label{tab:stats_purchases_per_user_training}
  \begin{tabular}{c|rrrrrrrr}
    \toprule
    Number of Purchases	per User & Mean		& Stddev			& Min	& Quan. 0.25	 &	Quan. 0.5 (Median)	& Quan. 0.75		& Quan. 0.99		& Max	\\
    \midrule
    $N$							& 97.31		& 25,949.54		& 1		& 25				& 58								& 116				& 511				& 115,288,138		\\
    \midrule
   $b$							& 10.72		& 1,719.44		& 1		& 2				& 6								& 13					& 65					& 7,634,957	 \\
  \bottomrule
\end{tabular}
\end{table*}

The pre-training data has the interaction (purchase) sequences of users
in the period from August 16 2018 to September 15 2019, with the cutoff time $T$ as the beginning of August 16 2019.
That is, $T - D_1$ is the beginning of August 16 2018,
and $T + D_2$ is the end of September 15 2019.
We randomly sampled 20 million users from hundreds of millions of users who had
at least one purchase both in $[T - D_1, T)$ and in $[T, T + D_2)$.
These 20 million interaction (purchase) sequences, one per each user,
involve totally 98.7 million items.

Table \ref{tab:stats_purchases_per_user_training} lists some statistics of the number of purchases per user both in $[T - D_1, T)$ and in $[T, T + D_2)$.
These statistics include the mean, standard devation, minimum, quantile 0.25, quantile 0.5, quantile 0.75, quantile 0.99 and maximum.
The table clearly shows the distribution of the number of purchases per user has a long tail.
Take $N$, the number of purchases per user within the period from August 16 2018 to August 15 2019, for example.
The mean of $N$ is 97.31, the median of $N$ is 58, the quantile 0.99 of $N$ reaches a big value 511.
We partition the whole dataset by users into training dataset (which has 18M users) and validation dataset (which has 2M users)
so that the hyper-parameters can be tuned based on the validation dataset.

The GUIM model for the pre-training has three transformer layers,
where the dimensions in the category-id embedding table, the item-id embedding table, and the (textual) word embedding table,
denoted by $d_c$, $d_i$, and $d_w$, are 128, 128 and 32 respectively.
In the baseline model with $C = 1$, the dimension of both a user embedding and an item embedding is also 128.
Accordingly, in the baseline model with $C = 1$,
the number of parameters in the three embedding tables, the three-layer transformer, and all other components are
43.84M, 0.59M, and 0.12M respectively.


\subsection{Downstream Evaluation Tasks}
We will introduce two sets of downstream evaluation tasks for our GUIM model.
For both evaluation-task sets,
we generate the interaction sequences from September 16 2018 to October 15 2019
but mask all the interaction information from September 16 2019 to October 15 2019 to avoid the possible information leak.
Then we feed these generated interaction sequences into our pre-trained GUIM model to obtain the relevant user embeddings and item embeddings.
Thus, we directly use the pre-trained GUIM model for inference (without fine-tuning our GUIM).


\subsubsection{Consumer Matching Prediction Tasks}\hspace*{\fill} \\

In this subsection, we introduce a set of Consumer Matching Prediction (CMP) tasks to evaluate the quality of general user and item representation, in terms of its capability of predicting both long-term and short-term interests of a user in various items.
We generate the datasets CMP-L, CMP-S, CMP-N, which address long-term, short-term and (immediate) next interests of a user respectively.

The dataset CMP-L is built to evaluate the ability of user and item embeddings to predict a user's interests over a relatively long period (30 days).
It includes 100,000 * 1,046,795 records, as a result of a Cartesian set with 100k users and roughly 1M items.
While the users are randomly sampled,
the items include both the top items across two websites (Taobao and Tmall)
as well as all the items that the sampled 100k users purchased in the next 30 days (in the period from September 16 2019 to October 15 2019) to ensure all positive items are included in the recall@M calculation (which will be explained in Eq. (\ref{eq_recall})).
If and only if a user purchases an item in the next 30 days, the label of the pair of the user and the item is positive.

The dataset CMP-S is built to evaluate the ability of user and item embeddings to predict user's relatively short-term interests on the next day.
It includes 165,262 * 1,016,995 records, as a result of a Cartesian set of roughly 165k users and 1M items.
While the users are randomly sampled,
the items include both the top items across two website
as well as all the items that the 165k users purchased on the next day (on September 16 2019) to ensure all positive items are included in recall@M calculation.
If and only if the user purchases the item in the next day, the label of the pair of the user and the item is positive.

Similarly, the dataset CMP-N is built to evaluate the ability of user and item embeddings to predict
the immediate next item that a user will purchase.
It includes 445,776 * 500,000 records, a Cartesian set of roughly 445k users and 500k items.
If and only if the item is the first item purchased by the user in the next 30 days (in the period from September 16 2019 to October 15 2019), the label of the pair of the user and the item is positive.

Note that we choose the aforementioned numbers of users and items in these three datasets
so that they are large enough to lead to the precise measurement
while keeping the corresponding computation costs reasonable.

Just as \cite{CTR_Wang:KDD2011,NSPR_Ebesu:IRJ2017,DropoutNet_Volkovs:NIPS2017},
we use top M recall as the evaluation metric.
For each user, the definition of recall@M is
\begin{equation}
\label{eq_recall}
\text{recall@M} = \frac{\text{number of items the user interacts with in top M}}{\text{total number of items the user interacts with}}.
\end{equation}
A higher recall@M indicates better prediction performance for the user.
The recall@M for the entire system can be summarized using the average of the recall@M of every user.
For all the CMP tasks we let M be 100, and we
call the aforementioned Alibaba's vector indexing engine to expedite the recall@100 computation.

\subsubsection{Consumer Profile Prediction Tasks}\hspace*{\fill} \\
In this subsection we introduce a set of Consumer Profile Prediction (CPP) tasks,
which is to predict the basic profile (characteristics) of a customer.
The CPP tasks are crucial for an e-commerce platform,
because successful consumer profile prediction serves as a basis
for marketing, promotions, and many other 
customer services.
In this paper, we choose the following three CPP tasks as our downstream tasks:
(1) "has child", which predicts whether a given user has a child or not.
(2) "baby gender", which predicts the gender of the child of a given user.
(3) "baby age group", which predicts the age level of of the child of a given user,
whose label value is in \{0, 1, ..., 6\}.
For each CPP task, we partition the labelled dataset into training set and test set.
There are roughly 10M records in the training set for each task,
and the sizes of three test sets are 100k, 200k, 100k respectively.
For each task, we obtain the general user embeddings from our pre-trained GUIM model,
and then feed each user embedding as the only features of the user into a three-layer fully connected feed-forward network (FFN)
for the training.
Thus, for our GUIM model, each user has totally $C \cdot d$ input features to train the FFN.
After the training, we get the prediction accuracy of the trained FFN on each test dataset.


\subsection{Evaluation Results}
In the GUIM, we utilize mixture of representation (MoR) as a novel way to represent users.
Now we will compare GUIM with other user-representation alternatives in terms of the performance in the downstream tasks.

\subsubsection{Comparison between GUIM and Its EDI-based Counterpart}\hspace*{\fill} \\
In this subsection, we compare the GUIM with a straightforward  baseline
which uses an embedding dimension increasing (EDI) approach in the general user and item representation learning.
We evaluate the GUIM and this baseline method (denoted by GUI-EDI) on the CPP and CMP tasks,
in order to see the difference between a user represented
with a set of vectors and that represented with a single vector of increased dimension.

More specifically,
in our GUIM, we represent each user with a set of $C$ vectors of dimension $d$ and represent each item by a single vector of dimension $d$;
in the GUI-EDI, we use a single vector of dimension $C \cdot d$ to represent each user and each item.
In the GUI-EDI method, we also correspondingly increase the intermediate size in the FFN sub-layer in each transformer layer
to keep BERT's convention that the intermediate size is four times as large as the embedding size.
We pre-train both GUIM models and GUI-EDI models
sufficiently many epochs
to ensure that they converge on the validation dataset.


\begin{table*}[t]
  \caption{Performance Comparison between GUI-EDI and GUIM in CMP \& CPP Tasks ($d=128$ and $C \in \{1, 2, 3, 4 \}$)}
  \label{tab:EDI Comparsion with MoR}
  \begin{tabular}{ll|cccc|cccc}
    \toprule
    \multicolumn{2}{l|}{Task} & \multicolumn{4}{c|}{GUI-EDI} &  \multicolumn{4}{c}{GUIM} \\
    & & $C$=1 & $C$=2 & $C$=3 & $C$=4 & $C$=1 & $C$=2 & $C$=3 & $C$=4  \\
    \midrule
    \multirow{6}*{CMP}
    & CMP-L         & 0.0759 & 0.0768 & 0.0775 & 0.0748 & 0.0759 & 0.0783 & 0.0791 & 0.0802  \\
    & CMP-L Impr. \% & 0.00\% & 1.27\% & 2.11\% & -1.43\% & 0.00\% & 3.26\% & 4.25\% & 5.68\%  \\
    & CMP-S & 0.0780 & 0.0798 & 0.0801 & 0.0752 & 0.0780 & 0.0814 & 0.0823 & 0.0840  \\
    & CMP-S Impr. \% & 0.00\% & 2.33\% & 2.67\% & -3.63\% & 0.00\% & 4.28\% & 5.53\% & 7.63\%  \\
    & CMP-N & 0.1608 & 0.1624 & 0.1627 & 0.1580 & 0.1608 & 0.1654 & 0.1672 & 0.1687  \\
    & CMP-N Impr. \% & 0.00\% & 1.00\% & 1.21\% & -1.76\% & 0.00\% & 2.83\% & 3.94\% & 4.93\%  \\
    \midrule
    \multirow{6}*{CPP}
    & Has child & 0.8230 & 0.8268 & 0.8248 & 0.8295 & 0.8230 & 0.8279 & 0.8310 & 0.8331 \\
    & Has child Impr. \% & 0.00\% & 0.46\% & 0.22\% & 0.80\% & 0.00\% & 0.60\% & 0.98\% & 1.24\% \\
    & Baby Gender & 0.8066 & 0.8188 & 0.8184 & 0.7973 & 0.8066 & 0.8211 & 0.8252 & 0.8317  \\
    & Baby Gender Impr. \% & 0.00\% & 1.51\% & 1.46\% & -1.16\% & 0.00\% & 1.80\% & 2.31\% & 3.11\%  \\
    & Baby Age & 0.6291 & 0.6424 & 0.6457 & 0.6288 & 0.6291 & 0.6506 & 0.6588 & 0.6659  \\
    & Baby Age Impr. \% & 0.00\% & 2.11\% & 2.64\% & -0.04\% & 0.00\% & 3.42\% & 4.72\% & 5.85\% \\
  \bottomrule
\end{tabular}
\end{table*}

We pre-train four pairs of models by setting $d = 128$
while increasing the aforementioned $C$ from 1 to 4 in both the GUI-EDI and the GUIM.
Note that the GUI-EDI and the GUIM are exactly the same when $C=1$.
Then we evaluate these models in both CMP tasks and CPP tasks in terms of recall@100 and (classification) prediction accuracy respectively.
Table \ref{tab:EDI Comparsion with MoR} shows the evaluation results of these models
as well as their performance improvement percentage from the model with $C=1$.
As clearly demonstrated in the table, for each $C > 1$
the GUIM achieves better performance than its GUI-EDI counterpart in each of six tasks.
It can also be clearly seen that the performance of our GUIM keeps increasing as $C$ increases across all the six tasks.
For example, recall@100 for the CMP-L task increases from 0.0759 to 0.0802 (with 5.68\% improvement) when $C$ increases from 1 to 4.
In contrast, the performance of the GUI-EDI does not have such an increasing pattern:
it first increases and then decreases when $C$ increases in five out of six tasks.
In fact, when $C$ increases,
the pre-training time (per epoch) of the GUI-EDI increases along with a slower convergence rate,
while that of the GUIM remains almost the same.
We have tried to pre-train the GUI-EDI model with $C=4$ with five times more epochs,
but find that its performance in a downstream task often is still unsatisfactory.

\begin{table*}[t]
  \caption{Comparison in Number of Parameters between GUI-EDI and GUIM  ($d = 128$ and $C \in \{1, 2, 3, 4 \}$)}
  \label{tab:parameters_related_to_C}
  \begin{tabular}{c|cccc|cccc}
    \toprule
    Model Components & \multicolumn{4}{c|}{GUI-EDI} &  \multicolumn{4}{c}{GUIM} \\
                                        & $C$=1     & $C$=2     & $C$=3         & $C$=4         & $C$=1     & $C$=2     & $C$=3     & $C$=4 \\
    \midrule
    Final projection matrix in $f_j()$  & 36,864    & 73,728    & 110,592       & 147,456       &  36,864	& 36,864	& 36,864	& 36,864 \\
    Embedding table in $f_t()$          & 46,848    & 93,696    & 140,544       & 187,392       &  46,848	& 46,848	& 46,848	& 46,848 \\
    Embedding table in $f_{cls}()$      & 128       & 256       & 384           & 512           &  128	    & 256	    & 384	    & 512 \\
    3-layer Transformer                 & 589,824   & 2,359,296 & 5,308,416     & 9,437,184     &  589,824	& 589,824	& 589,824	& 589,824 \\
    Projection matrix in $f_u()$        & 16,384    & 65,536	& 147,456	    & 262,144       &  16,384	& 16,384	& 16,384	& 16,384 \\
    Projection matrix in $f_o()$        & 16,384    & 65,536	& 147,456	    & 262,144       &  16,384	& 16,384	& 16,384	& 16,384 \\
    \midrule
    Summation of the above              & 706,432   & 2,658,048 & 5,854,848     & 10,296,832    & 706,432	& 706,560	& 706,688	& 706,816 \\
    \bottomrule
\end{tabular}
\end{table*}

Also note that our experimental setting actually favors the GUI-EDI, because the GUI-EDI has a larger number of parameters than
the GUIM counterpart when $C > 1$.
Table \ref{tab:parameters_related_to_C} shows the difference in the number of parameters
between the components in the GUI-EDI and the ones in the GUIM when $C$ increases.\footnote{The item-id embedding table,
the category-id embedding table, and the (textual) word embedding table, have their sizes invariant to $C$
in both the GUI-EDI and the GUIM, because the corresponding dimensions $d_i$, $d_c$ and $d_w$ are invariant to $C$.
Actually, the number of parameters in the item-id embedding table, the category-id embedding table, and the word embedding table
are always 25.60M, 1.28M and 16.96M respectively when $d = 128$.}
Some model components in the GUI-EDI,
including each transformer layer, the output projection matrices in $f_u()$ and $f_o()$ in Figure \ref{fig_GUIM_model},
have their number of parameters quadratic with respect to $C$;
while other components in the GUI-EDI,
including the time embedding table, the CLS token embedding table, and the final projection matrix in $f_j()$,
have their number of parameters linear with respect to $C$.
In contrast,
in the GUIM, only one small component, the CLS token embedding table, has its size linear with respect to $C$,
and all the other components shown in Table \ref{tab:parameters_related_to_C} have their sizes invariant to the increase in $C$.
As shown in Table \ref{tab:EDI Comparsion with MoR}, however, the GUIM still outperforms its GUI-EDI counterpart when $C > 1$ even if
its size is smaller and its pre-training time is much shorter.


\subsubsection{Comparison between Different MoR Methods for User Representation}\hspace*{\fill} \\

In this section, we compare the GUIM with an alternative, denoted by GUIM-$MH$, to construct a user embedding as a set of vectors.
In the GUIM-$MH$, we put an additional self-attention layer on top of the final transformer output layer in the GUIM model with $C=1$,
and then construct a user embedding from the multi-head outputs of this additional layer.
We use $H$ to denote the number of heads in this additional self-attention layer of the GUIM-$MH$.
Different from the original setting in the transformer \citep{Transformer_Vaswani:NIPS2017},
in this additional self-attention layer, we set the size of projection matrices for query, key and value to be $d \times d$, instead of $d \times (d / H)$
so that the dimension of each head is $d$.
We then use these $H$ vectors of dimension $d$ to represent a user in the GUIM-$MH$.
When $H$ equals to $C$, both the GUIM and the GUIM-$MH$ use the same number of parameters to represent a user.

\begin{figure*}[htbp]
  \centering
  \includegraphics[width=\linewidth]{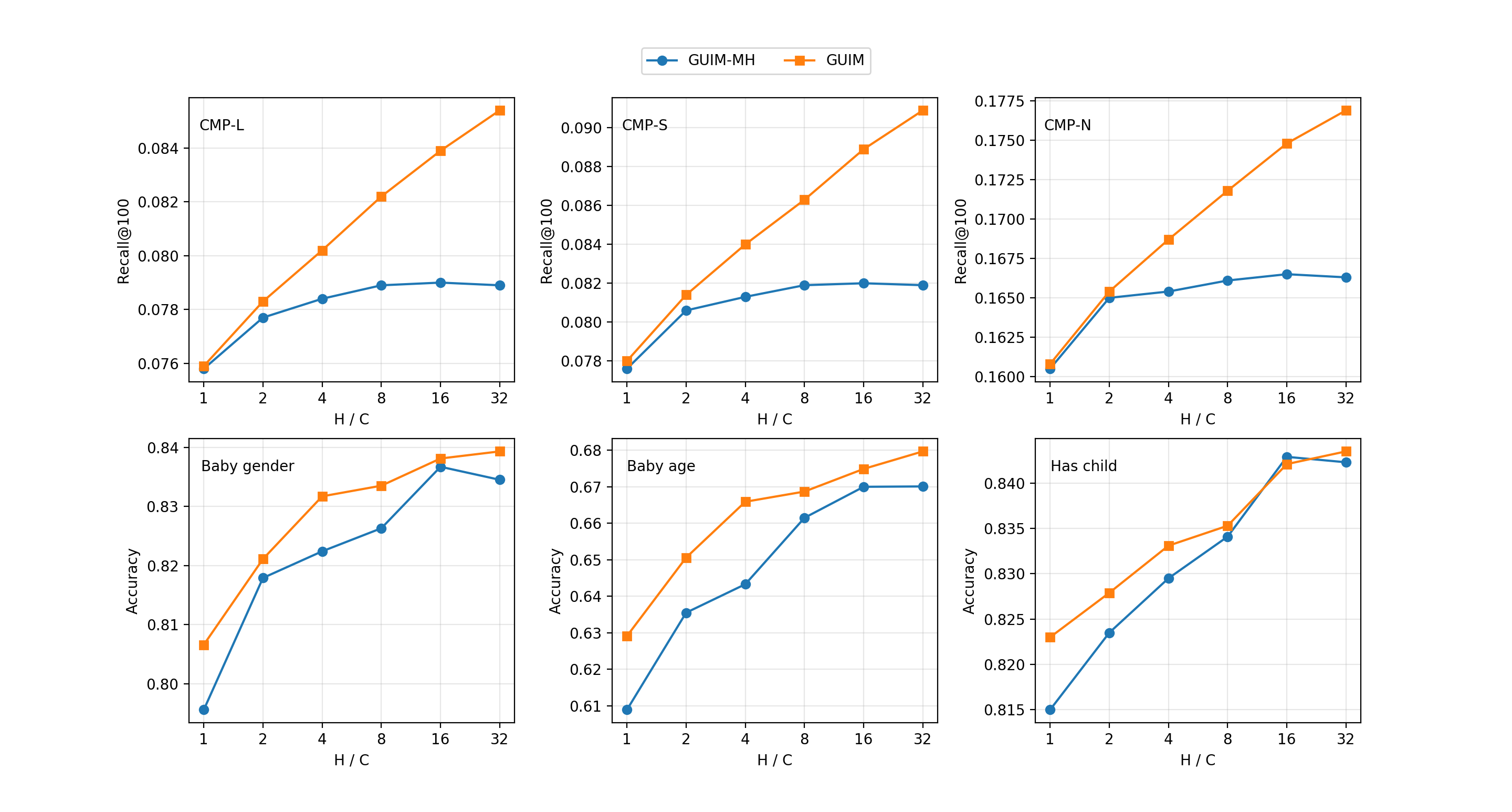}
  \caption{Performance Comparison between GUIM and GUIM-$MH$ on CMP \& CPP Tasks ($d=128$)}
  \label{fig_MoRvsMH_model}
\end{figure*}

\begin{figure*}[hp]
  \centering
  \includegraphics[width=\linewidth]{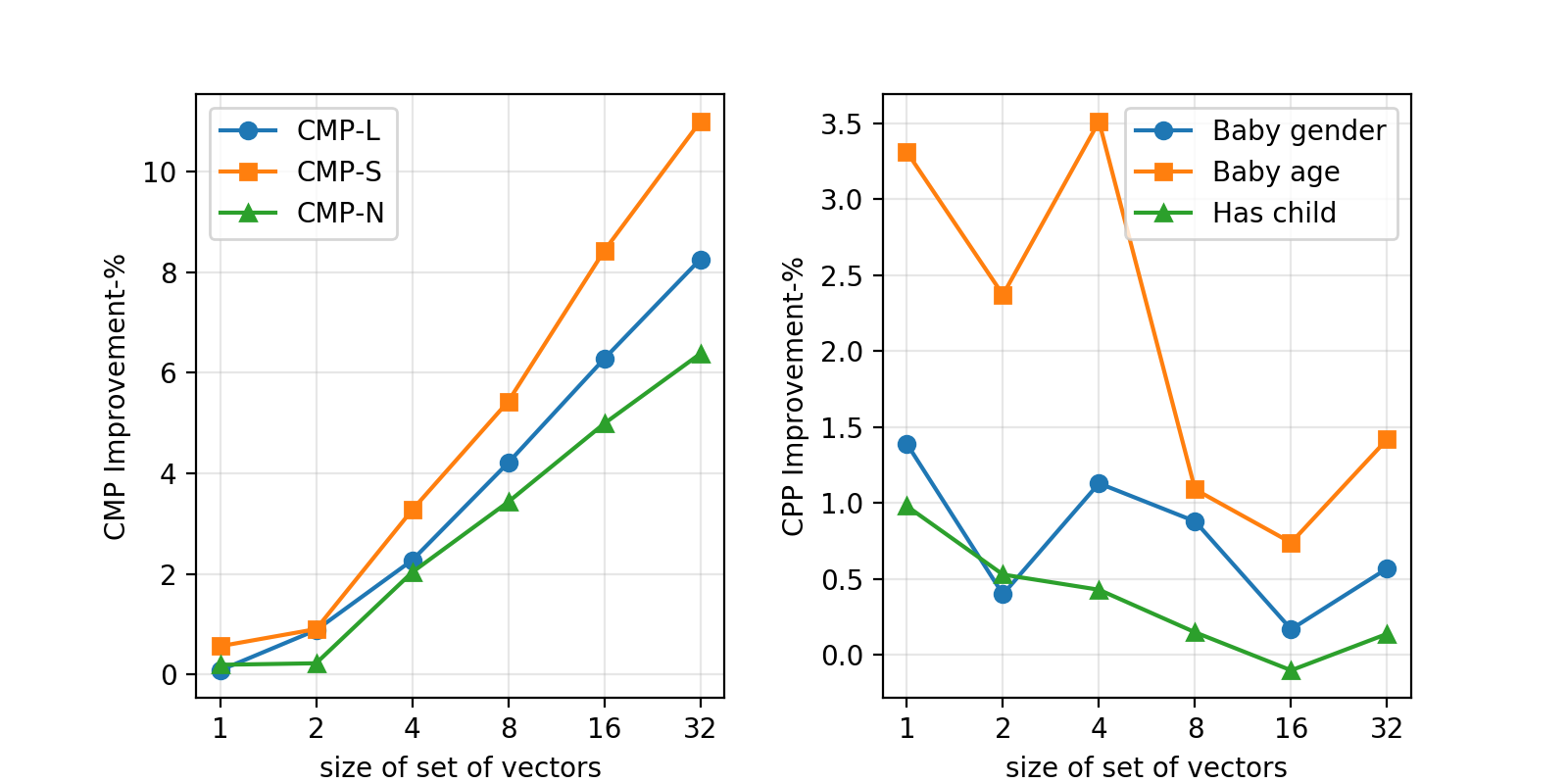}
  \caption{Improvement Percentage of GUIM over GUIM-$MH$ with Different Number of Vectors for User Representation on CMP \& CPP Tasks}
  \label{fig_ImprovMoRvsMH_model}
\end{figure*}

Figure \ref{fig_MoRvsMH_model} shows performance comparison between the GUIM and the GUIM-$MH$
when $H$ (or $C$) increases exponentially from 1 to 32 and $d$ is fixed at $128$.
It can be clearly seen that the performance of the GUIM is almost universally better than that of the GUIM-$MH$
when the number of vectors for user representation ($H$ or $C$) is larger than 1 in all the six investigated tasks.
While the advantages of the GUIM over the GUIM-$MH$ exist in all these six tasks,
such advantages are particularly large in three CMP tasks.
From Figure \ref{fig_MoRvsMH_model},
we can see that
the increase in recall@100 from the GUIM with respect to the number of vectors is nearly log-linearly,
while that from the GUIM-$MH$ shows much faster saturation.
We also show the relative improvement percentage of the GUIM over the GUIM-$MH$ in Figure \ref{fig_ImprovMoRvsMH_model}.
When the number of vectors equals $32$, the relative improvement percentage of the GUIM over the GUIM-$MH$ are 8.26\%, 11.0\%, 6.38\% in CMP-L, CMP-S, CMP-N tasks respectively.





\section{CONCLUSIONS and FUTURE WORK \label{sec_conclusions}}
In this paper we propose the GUIM model to learn general representation (embeddings) for users and items,
which is pre-trained through the massive user-item interaction data, 
and then can be used for various downstream tasks across platforms in Alibaba e-commerce ecosystem.
To better model users' diverse interests,
we use a mixture-of-representation (MoR) strategy as a novel way to represent each user as a set of vectors.
We introduce a matching task along with the MLM task proposed in the BERT to pre-train our GUIM.
Additionally, we use the InfoNCE during the pre-training to solve the computational challenge
coming from a huge (item) vocabulary size.
The experimental results on large-scale real datasets clearly show the value of our model.

In the future,
we will study the benefits when more action types (such as clicking, browsing and searching) are added into a user-item interaction sequence.
We want to see whether the added information can provide a learned embedding with some extra insights about the characteristics of a user such as his/her shopping pattern.
We also plan to combine more information modes (such as images, videos and user reviews) of an item in the model input to see the potential benefits.
Additionally, we will expand a user-item interaction sequence across multiple years to see whether our model can learn more patterns such as seasonality.
Finally, we will try to design new pre-training tasks which are able to further help the representation learning,
especially for extremely long interaction sequences.

\bibliographystyle{ACM-Reference-Format}
\bibliography{GUIM}


\end{document}